# Sensitivity Analysis for Threshold Decision Making with Dynamic Networks

**Theodore Charitos** and **Linda C. van der Gaag**
Department of Information and Computing Sciences, Utrecht University
P.O Box 80.089, 3508 TB Utrecht, The Netherlands
{theodore,linda}@cs.uu.nl

## Abstract

The effect of inaccuracies in the parameters of a dynamic Bayesian network can be investigated by subjecting the network to a sensitivity analysis. Having detailed the sensitivity functions involved in our previous work, we now study the effect of parameter inaccuracies on a recommended decision in view of a threshold decision-making model. We describe the effect of varying one or more parameters from a conditional probability table and present a computational procedure for establishing bounds between which assessments for these parameters can be varied without inducing a change in the recommended decision. We illustrate the various concepts by means of a real-life dynamic network in the field of infectious disease.

## 1 Introduction

Probabilistic graphical models are often used in contexts where human decision makers have to make a decision in uncertainty. The marginal probability distributions yielded by the model then are taken as input to a decision-making model. The simplest model for choosing between alternative decisions is the threshold decision-making model, in which an output probability is compared against a number of fixed threshold probabilities which demarcate the boundaries for the various decisions [13]. In our application for ICU care, for example, a clinician has to decide whether or not to start antibiotics treatment for a patient who is suspected of having ventilator-associated pneumonia (VAP), based upon the probability of VAP being present.

Probabilistic graphical models are typically learned from data or constructed with the help of domain experts. Due to incompleteness of data and partial knowledge of the domain under study, the numerical parameters of the model tend to be inaccurate to at least some degree. The inaccuracies may affect the output probabilities of the model as well as the decisions based upon these probabilities. The effects of inaccuracies in the parameters of a network on its output probabilities can be studied by subjecting the network to a *sensitivity analysis* [2, 7, 8, 9]. In view of a decision-making model, however, robustness of the output of a probabilistic graphical model pertains not just to the computed output probabilities but also to the decisions based upon these probabilities. In this paper, we study this type of robustness for dynamic Bayesian networks (DBNs) in view of the threshold decision-making model.

Previous work on sensitivity analysis of Bayesian networks (BNs) in general showed that any posterior probability for an output variable is a quotient of two linear functions in any of the network's parameters [8]; the posterior probability can further be expressed as a sum of such functions in all parameters from a single conditional probability table [2]. Building upon these results, we show that in sensitivity analysis of DBNs a quotient of polynomial functions is obtained, where the order of these polynomials is linear in the time scope taken into consideration [4, 6]. We further show how the resulting functions can be used to study the robustness of a decision that is based upon an output probability of the network. By doing so, we focus not just on the effect of varying a single parameter, but also on the effect of varying all parameters from a given conditional probability table. In our medical application, for example, we thus provide for studying the extent to which the sensitivity and specificity rates of a particular diagnostic test can be varied without affecting the clinician's decision for treatment. We illustrate the various concepts involved by means of a sensitivity analysis of our dynamic network in the field of infectious disease [5].

Establishing the sensitivity functions for a DBN has a time complexity similar to that of performing exact inference in such a model. Using quotients of higher-order polynomials for further processing in view of the threshold decision-making model can also be highly demanding from a computational point of view [6]. To handle the complexity involved, we present an approximate method for sensitivity analysis that reduces the runtime requirements involved yet incurs only a small loss in accuracy.

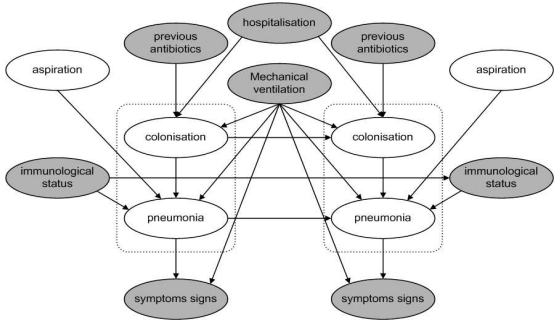

Figure 1: The dVAP model for the diagnosis of VAP for two consecutive time steps; clear nodes are hidden, shaded nodes are observable. The dashed boxes indicate the hidden processes.

## 2 Preliminaries

The simplest type of dynamic network is a hidden Markov model (HMM) $H = (X, Y, A, O, \Gamma)$ involving a single stochastic process [14]. We use $X_n$ to denote the hidden variable of the modelled process at time step $n$. $X_n$ has the possible states $x_i^n$, $i = 1, \ldots, l$, $l \geq 2$. The *transition matrix* for $X_n$ is denoted as $A = \{a_{i,j}\}$ with elements $a_{i,j} = p(X_{n+1} = x_j^{n+1} \mid X_n = x_i^n)$, $i, j = 1, \ldots, l$, for all $n$. We denote the observable variables by $Y_n$, with values $y_j^n$, $j = 1, \ldots, m$, $m \geq 2$. The value of $Y_n$ is generated from the state of the hidden variable according to a time-invariant *observation matrix* $O = \{o_{i,j}\}$ with $o_{i,j} = p(Y_n = y_j^n \mid X_n = x_i^n)$, $i = 1, \ldots, l$, $j = 1, \ldots, m$, for all $n$. We further denote by $\Gamma = \{\gamma_i\}$ the *initial probability vector* for the hidden variable, with $\gamma_i = p(X_1 = x_i^1)$, $i = 1, \ldots, l$. A DBN of more general structure is an extension of an HMM, capturing a compound process that involves a collection of hidden variables. We assume that our dynamic networks are time invariant, that is, neither the topology nor the parameters of the model change across time steps.

In this paper, we focus on the task of *monitoring* in DBNs, that is, on computing marginal distributions for the model's hidden variables for some time step $n$ given the observations that are available up to and including that time step. For this purpose, the *interface algorithm* is available [11], which basically is an extension of the junction-tree algorithm for probabilistic inference in graphical models in general. The interface algorithm efficiently exploits the concept of *forward interface*, which is the set of variables at time step $n$ that affect some variables at time step $n+1$ directly; in the sequel, we use $\boldsymbol{FI}$ to denote this set of variables. The computational complexity of the interface algorithm is exponential in the number of hidden variables at each time step, which is prohibitive for larger models.

Throughout the paper we will use the *dVAP network* for illustration purposes. The dVAP network is a real-life DBN for diagnosing VAP in ICU patients and is destined for use in clinical practice [5]. The network includes two interacting hidden processes (*colonisation* and *pneumonia*), three input processes (summarised in *immunological status*), three input observable variables (*hospitalisation*, *mechanical ventilation*, and *previous antibiotics*), one hidden input variable (*aspiration*), and seven output observable variables (summarised in *symptoms-signs*). Per time step, representing a single day, the model includes 30 variables. Each of the interacting processes consists of seven subprocesses that are a-priori independent. The transition matrices of these subprocesses are moderately stochastic. Figure 1 shows the dVAP network in a compact way.

## 3 Sensitivity analysis revisited

Having been studied extensively in the context of BNs, sensitivity analysis has received less attention in DBNs. In this section, we briefly review previous work on sensitivity analysis in BNs and extend it to a dynamic context.

### 3.1 Sensitivity analysis of BNs

Sensitivity analysis of a BN amounts to establishing, for each of the network's parameter probabilities, a function that expresses an output probability of interest in terms of the parameter under study [7, 9, 16]. We take the posterior probability $p(b \mid e)$ for our probability of interest, where $b$ is a specific value of the variable $B$ and $e$ denotes the available evidence; we further let $\theta = p(h_i \mid \pi)$ be our parameter under study, where $h_i$ is a value of the variable $H$ and $\pi$ is a specific combination of values for the parents of $H$. The sensitivity of the probability $p(b \mid e)$ to variation of the parameter $\theta$ now is expressed by a *sensitivity function* $p(b \mid e)(\theta)$. If the parameters $p(h_j \mid \pi)$, $h_j \neq h_i$, specified for $H$ are co-varied proportionally according to

$$p(h_j \mid \pi)(\theta) = \begin{cases} \theta & \text{if } j = i \\ p(h_j \mid \pi) \cdot \frac{1-\theta}{1-p(h_i \mid \pi)} & \text{otherwise} \end{cases}$$

for $p(h_i \mid \pi) < 1$, then this function is a quotient of two linear functions in $\theta$, that is,

$$p(b \mid e)(\theta) = \frac{p(b, e)(\theta)}{p(e)(\theta)} = \frac{c_1 \cdot \theta + c_0}{d_1 \cdot \theta + d_0}$$

where $c_1, c_0, d_1$ and $d_0$ are constants with respect to $\theta$ [7]. Under the assumption of proportional co-variation, therefore, any sensitivity function is characterised by at most three constants. Note that for parameters of which the probability of interest is algebraically independent, the function simply equals the posterior probability $p(b \mid e)$. Any computations can therefore be restricted to the *sensitivity set* for the variable of interest, which can be readily established from the network's graphical structure. An efficient scheme for sensitivity analysis is available [8], which requires an inward propagation in the junction-tree for processing evidence and an outward propagation for establishing the constants of the sensitivity functions for all parameters per output probability.

## 3.2 Sensitivity analysis of DBNs

The main difference with sensitivity analysis of BNs is that a parameter occurs multiple times in a DBN. In previous work [4], we derived functional forms to express the sensitivity of a probability of interest of an HMM in terms of a parameter under study. We briefly review these sensitivity functions. We begin by studying the sensitivities of an HMM in which no evidence has been entered as yet. The probability of interest is the prior probability $p(x_r^n)$ of some state $x_r$ of the hidden variable $X_n$. Let $\theta_a = a_{i,j} \in A$ be a transition parameter in the model such that $X_n$ is algebraically dependent on $\theta_a$. Then,

$$p(x_r^n)(\theta_a) = c_{n,r}^{n-1} \cdot \theta_a^{n-1} + \ldots + c_{n,r}^1 \cdot \theta_a + c_{n,r}^0$$

where $c_{n,r}^{n-1}, \ldots, c_{n,r}^0$ are constants with respect to $\theta_a$ dependent on $n$; note that the function is a polynomial of order $n-1$ in the parameter under study. For an initial parameter $\theta_\gamma = \gamma_i \in \Gamma$, the function is linear

$$p(x_r^n)(\theta_\gamma) = c_{n,r}^1 \cdot \theta_\gamma + c_{n,r}^0$$

where $c_{n,r}^1$ and $c_{n,r}^0$ are constants with respect to $\theta_\gamma$. For an HMM in which no evidence has been entered, the observable variables do not belong to the sensitivity set of the hidden variable. Its prior probability therefore is algebraically independent of any observation parameter.

We now assume that some evidence has been entered into the model; we use $e_n$ to denote the cumulated evidence up to and including time step $n$. For the sensitivity function that expresses the posterior probability $p(x_r^n \mid e_n)$ in terms of a transition parameter $\theta_a = a_{i,j} \in A$, we find that

$$\frac{p(x_r^n, e_n)(\theta_a)}{p(e_n)(\theta_a)} = \frac{c_{n,r}^{n-1} \cdot \theta_a^{n-1} + \ldots + c_{n,r}^1 \cdot \theta_a + c_{n,r}^0}{d_{n,r}^{n-1} \cdot \theta_a^{n-1} + \ldots + d_{n,r}^1 \cdot \theta_a + d_{n,r}^0}$$

where $c_{n,r}^{n-1}, \ldots, c_{n,r}^0, d_{n,r}^{n-1}, \ldots, d_{n,r}^0$ again are constants with respect to $\theta_a$. The function thus is a quotient of two polynomials of order $n-1$. For an observation parameter $\theta_o = o_{i,j}$, the sensitivity function is

$$\frac{p(x_r^n, e_n)(\theta_o)}{p(e_n)(\theta_o)} = \frac{c_{n,r}^b \cdot \theta_o^b + \ldots + c_{n,r}^1 \cdot \theta_o + c_{n,r}^0}{d_{n,r}^n \cdot \theta_o^n + \ldots + d_{n,r}^1 \cdot \theta_o + d_{n,r}^0}$$

where $b = n$ if $r = i$ and $b = n - 1$ otherwise; $c_{n,r}^b, \ldots, c_{n,r}^0, d_{n,r}^n, \ldots, d_{n,r}^0$ are constants with respect to $\theta_o$. The order of the polynomials involved again grows linearly with $n$. For an initial parameter $\theta_\gamma$, to conclude, we have that the sensitivity function is a quotient of two linear functions. Similar results hold for probabilities of interest belonging to any possible time step $n_o < n$ or $n_o > n$ [4].

The results for HMMs are readily generalised to DBNs. We consider the posterior probability of interest $p(b_r^n \mid e_n)$ of the state $b_r$ of the hidden variable $B_n$ given the evidence $e_n$. Then, for any variable $H_n \in Sens(B_n, e_n)$, the sensitivity function expressing $p(b_r^n \mid e_n)$ in $\theta = p(h_i^n \mid \pi)$ is a quotient

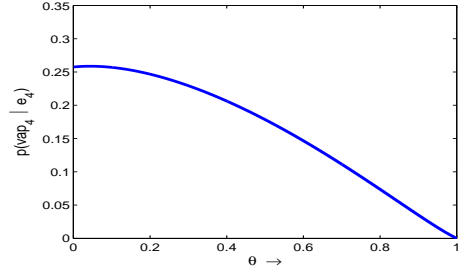

Figure 2: The sensitivity function expressing the probability of pneumonia at day 4 given evidence $e_4$ for a patient, in terms of the parameter $\theta = p(leucocytosis = yes \mid pneumonia = yes)$.

of two polynomials of order $n-1$ if $H_n \in \boldsymbol{FI}$, or of order $n$ otherwise.

As an example sensitivity function, Figure 2 depicts, for the dVAP network, the effect of varying the parameter $\theta = p(leucocytosis = yes \mid pneumonia = yes)$ on the probability of pneumonia at day 4 given evidence $e_4$ for a specific patient. The depicted function is a quotient of two polynomials of order 4 each. For computing the constants in the various sensitivity functions, we combined the interface algorithm with the junction-tree scheme for sensitivity analysis; further details of the resulting algorithm are out of the scope of this paper.

## 4 Threshold decision making

BNs in general yield marginal probability distributions for their output. Often these marginal distributions are input to a decision maker who has to make a decision. The simplest model for choosing between alternative decisions is the threshold decision-making model. In this section, we briefly review the threshold model for decision making and describe sensitivity analysis of BNs in view of this model.

Although generally applicable, the threshold decision-making model is used most notably for patient management in medicine [13]. Since our dVAP network also pertains to the field of medicine, we present the model in medical terms. With the model, a clinician decides whether or not to gather additional information from diagnostic tests and whether or not to give treatment based upon the probability of disease for a patient. The threshold model to this end builds upon various threshold probabilities of disease.

The treatment threshold probability $p^*$ is the probability at which the clinician is indifferent between giving treatment and withholding treatment. If, for a specific patient, the probability of disease exceeds the treatment threshold probability, then the clinician will decide to treat the patient as if the disease were known to be present with certainty. Alternatively, if the probability of disease is less than $p^*$, the clinician will basically withhold treatment. As a consequence of the uncertainty concerning the true condition of the patient however, additional information from a diagnostic test may affect the clinician's basic management decision. The threshold model therefore includes another two threshold probabilities. The no treatment-test threshold probability of disease $p^-$ is the probability at which the clinician is indifferent between the decision to withhold treatment and the decision to obtain additional diagnostic information. The test-treatment threshold probability $p^+$ is the probability at which the clinician is indifferent between obtaining additional information and starting treatment rightaway. Figure 3 illustrates the various threshold probabilities employed by the model.

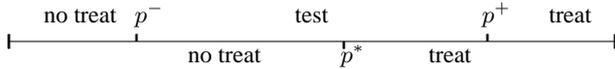

Figure 3: The threshold decision model.

In view of the threshold model for decision making, sensitivity of the output of a network pertains no longer to just a probability of interest computed from the network, but also to the decision based upon it. To take the various threshold probabilities employed into consideration, the method of sensitivity analysis of BNs has been enhanced with the computation of upper and lower bounds between which a network's parameters can be varied without inducing a change in decision [15]. The computation of these bounds builds upon the sensitivity functions relating the probability of interest to the network's parameters. By equating the function for a specific parameter to the various threshold probabilities, bounds are obtained between which the parameter can be varied. Since the sensitivity functions for a BN are either monotonically non-decreasing or monotonically non-increasing, a single lower bound and a single upper bound are guaranteed to exist.

## 5 Sensitivity analysis of decisions with DBNs

The probabilities established from a dynamic network are also often employed for decision making. The goal of the dVAP model, for example, is to monitor the onset of ventilator-associated pneumonia in ICU patients and to start appropriate treatment as soon as possible. Sensitivity to parameter variation then pertains not just to the probability of VAP but also to the decision that the clinician makes based upon this probability. To provide for studying this type of sensitivity, we extend sensitivity analysis of DBNs in view of threshold decision making.

### 5.1 Analysis of single parameters

Suppose that a posterior probability $p(x_r^n \mid e_n)$ of interest has been computed from a DBN; based upon this probability, a particular decision has been established from the threshold decision-making model. We now are interested in the effect of variation of a parameter $\theta$ on this decision. To compute upper and lower bounds between which the parameter can be varied without inducing a change in decision, the sensitivity function $p(x_r^n \mid e_n)(\theta)$ is equated to the threshold probabilities $p^-$ and $p^+$, respectively. The lower and upper bounds thus are solutions of the equations

$$\frac{p(x_r^n, e_n)(\theta)}{p(e_n)(\theta)} = p^- \quad \text{and} \quad \frac{p(x_r^n, e_n)(\theta)}{p(e_n)(\theta)} = p^+ \quad (1)$$

We recall that for DBNs a sensitivity function in general is a quotient of higher-order polynomials. Contrary to threshold decision making for BNs therefore, there is no guarantee that these functions are monotonically non-decreasing or non-increasing. The equations stated above may thus have multiple solutions instead of single ones.

We begin by studying a parameter for which single solutions exist for the two threshold equations above. Suppose that the lower and upper bounds computed from the equations are $\theta^-$ and $\theta^+$ respectively. If $p(x_r^n \mid e_n) < p^-$, then the decision to withhold treatment remains unaltered for any value of $\theta$ within the interval $(-\infty, \theta^-) \cap [0, 1]$. If $p(x_r^n \mid e_n) > p^+$, the decision to start treatment immediately remains unaltered for any value of $\theta$ within $(\theta^+, +\infty) \cap [0, 1]$. If $p^- \leq p(x_r^n \mid e_n) \leq p^+$, then the decision to gather further information will be the same for any value of $\theta$ within the interval $(\theta^-, \theta^+) \cap [0, 1]$.

As an example from the dVAP network, we illustrate the bounds on variation of the parameter $\theta = p(leucocytosis = yes \mid pneumonia = yes)$ in view of the management decision for a particular patient at day 4; Figure 2 shows the sensitivity function that expresses the probability of VAP for this patient in terms of $\theta$. Suppose that the three threshold probabilities are fixed at $p^* = 0.2, p^- = 0.12$, and $p^+ = 0.64$. From the dVAP network, we have that $p(vap_4 \mid e_4) = 0.134$ and hence that $p^- \leq p(vap_4 \mid e_4) \leq p^+$. The clinician thus decides to gather additional information for the patient. Solving the two threshold equations from (1), we find a lower bound on the parameter under study equal to $\theta^- = 0.676$; the upper bound is $\theta^+ = 1.1063$. For any value of the parameter within the interval $(0.676, 1]$, therefore, the decision to gather additional information will remain unaltered. Since the original value of the parameter has been assessed at $0.7$, we conclude that the decision is not very robust with regard to this parameter.

We now turn to parameters for which the threshold equations have multiple solutions. Suppose that from the no-treatment-test threshold probability $p^-$, we find the vector of solutions $\boldsymbol{\theta}^- = (\theta_1^-, \theta_2^-, \ldots, \theta_r^-)$, in which the parameter values $\theta_i^-$ are given in ascending order; from the test-treatment threshold probability $p^+$, we find the vector $\boldsymbol{\theta}^+ = (\theta_1^+, \theta_2^+, \ldots, \theta_s^+)$, again with the $\theta_i^+$ in ascending order. We further use $v(\theta_i)$ to denote the value of the first derivative of the sensitivity function for the parameter $\theta$ at $\theta_i$. The value of the first derivative helps in determining the intervals in which a particular decision holds. Now, if the output probability $p$ is smaller than $p^-$, the decision to withhold treatment remains unaltered for any value of $\theta$ that belongs to the compound interval

$$\Theta_+^- = \begin{cases} [0, \theta_1^-) \cup (\theta_2^-, \theta_3^-) \cup \ldots \cup (\theta_r^-, 1] & r \text{ is even} \\ [0, \theta_1^-) \cup (\theta_2^-, \theta_3^-) \cup \ldots \cup (\theta_{r-1}^-, \theta_r^-) & r \text{ is odd} \end{cases}$$

whenever $v(\theta_1^-) > 0$, and for any value of $\theta$ belonging to

$$\Theta_-^- = \begin{cases} (\theta_1^-, \theta_2^-) \cup (\theta_3^-, \theta_4^-) \cup \ldots \cup (\theta_{r-1}^-, \theta_r^-] & r \text{ is even} \\ (\theta_1^-, \theta_2^-) \cup (\theta_3^-, \theta_4^-) \cup \ldots \cup (\theta_r^-, 1] & r \text{ is odd} \end{cases}$$

whenever $v(\theta_1^-) < 0$. Similarly, if the output probability $p$ is larger than $p^+$, we find compound intervals $\Theta_-^+$ and $\Theta_+^+$ for the parameter $\theta$ within which the decision to start treatment immediately remains unaltered. Finally, if the output probability lies between $p^-$ and $p^+$, the vectors $\boldsymbol{\theta}^-$ and $\boldsymbol{\theta}^+$ are merged into the vector $\boldsymbol{\theta}^m = (\theta_1^m, \theta_2^m, \ldots, \theta_q^m)$, $q = r + s$. Now, the decision will be the same for any value of $\theta$ within the interval $\Theta^m$

$$\Theta^m = \begin{cases} [0, \theta_1^m) \cup (\theta_2^m, \theta_3^m) \cup \ldots \cup (\theta_{q-1}^m, \theta_q^m) & v(\theta_1^m) < 0 \\ (\theta_1^m, \theta_2^m) \cup (\theta_3^m, \theta_4^m) \cup \ldots \cup (\theta_{q-1}^m, \theta_q^m) & v(\theta_1^m) > 0 \end{cases}$$

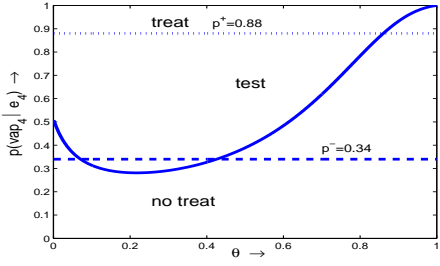

Figure 4: Threshold decision making for the treatment of pneumonia when varying the parameter $\theta = p(temperature = low \mid pneumonia = no, a.drugs = yes)$.

Note that in case $v(\theta_1^m) > 0$ and the value of the sensitivity function for $\theta = 0$ is greater than $p^-$, the interval $\Theta^m$ is the same as when $v(\theta_1^m) < 0$.

As an example, Figure 4 depicts the probability of pneumonia at day 4 given evidence $e_4$ in terms of the parameter $\theta = p(temperature = low \mid pneumonia = no, a.drugs = yes)$; the original value of $\theta$ equals 0.2, giving $p(vap_4 \mid e_4) = 0.278$. The figure also shows the intersection points with the threshold probabilities, which have been set at $p^- = 0.34$ and $p^+ = 0.88$. We compute the two lower bounds to be $\theta_1^- = 0.0918$ and $\theta_2^- = 0.4335$; the upper bound is $\theta_1^+ = 0.8781$. Using $v(\theta_1^-) < 0$, we find that the decision to withhold treatment remains unaltered for any value of $\theta$ in $(0.0918, 0.4335)$. We conclude that the decision is relatively robust with regard to the parameter.

### 5.2 Analysis of full conditional probability tables

In addition to single parameters, we may be interested in the effects of varying multiple parameters, for example from a single conditional probability table (CPT) [3]. In a medical application for instance, we may wish to study the robustness of a decision in terms of both the sensitivity and the specificity of a particular diagnostic test and not just in one of these rates. Recall that these rates express the probabilities that a test result is found to be positive (negative) in a patient who does (does not) have the disease. We now extend the previous results for single parameters to CPTs to provide for such an analysis.

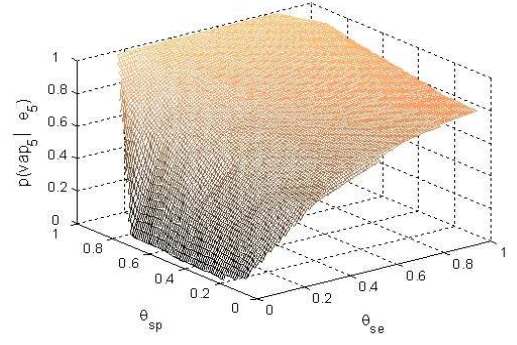

Figure 5: The sensitivity function expressing the probability of pneumonia given $e_5$ in terms of the sensitivity $\theta_{se}$ and specificity $\theta_{sp}$ rates of the CPT for radiological signs.

For BNs, any posterior probability for the output variable is a quotient of sums of linear functions in the parameters of a CPT [3, 8]. For a dynamic network we obtain sums of polynomial functions. For a joint probability $p(b_r^n, e_n)$ we find that

$$p(b_r^n, e_n) = \sum_{i=1}^{|\pi_{C_n}|} g_i(\theta_i)$$

where $|\pi_{C_n}|$ denotes the number of parent configurations of the variable $C_n$ and $g_i(\theta_i)$ represents a polynomial function in the parameter $\theta_i$ for a specific parent configuration for $C_n$. The polynomial functions $g_i(\theta_i)$ are all of the same order and can be computed individually using the considerations of the previous section. For the joint probability $p(e_n)$ a similar result holds. We conclude that, for a DBN, the sensitivity function for a CPT is a quotient of sums of higher-order polynomials in the parameters under study.

As an example, Figure 5 illustrates the effect of varying the sensitivity and specificity rates of the CPT for radiological signs of pneumonia at day 5 given evidence $e_5$ for a specific patient. The depicted sensitivity function is a quotient of sums of two polynomial functions of order 5 each.

Upon studying the effects of varying all parameters from a CPT in view of threshold decision making, we have to solve threshold equations similar to (1). For a single parameter, we identified intervals for the parameter's value within which a decision remains unaltered. For a CPT, we now identify areas in higher-space with the same meaning. In the remainder of this section, we consider a CPT with sensitivity and specificity rates as in the previous example; similar results hold for more complex CPTs.

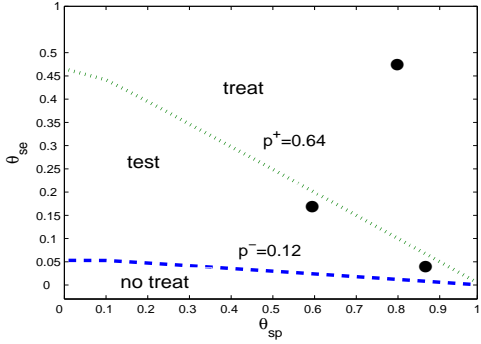

Figure 6: Threshold decision making when varying the sensitivity and specificity rates $\theta_{se}, \theta_{sp}$ of the CPT for radiological signs.

We begin again by studying a CPT for which single lower bounds exist for the two rates, denoted as $\theta_{se}^-$ and $\theta_{sp}^-$ respectively. By re-arranging the individual polynomial functions included, we can express the relationship between $\theta_{se}^-$ and $\theta_{sp}^-$ as

$$\widetilde{g}(\theta_{se}^-) = \widehat{g}(\theta_{sp}^-)$$

where $\widetilde{g}$ and $\widehat{g}$ are polynomials of the same order. If $\widetilde{g}$ is invertible in $[0, 1]$, we have that

$$\theta_{se}^- = \widetilde{g}^{-1}(\widehat{g}(\theta_{sp}^-))$$

which defines the relationship between the $\theta_{se}^-$ and $\theta_{sp}^-$. The *horizontal line test* can be used for checking whether $\widetilde{g}$ is invertible in $[0, 1]$. Establishing $\widetilde{g}^{-1}$, however, is computationally expensive if not infeasible [10].

To determine the relationship between $\theta_{se}^-$ and $\theta_{sp}^-$ and thereby study the robustness of a recommended decision, we use a numerical approximation procedure. We repeatedly pick a value $\theta_{se}'^- \in [0, 1]$ and solve for $\theta_{sp}'^- \in [0, 1]$. From the pairs $(\theta_{se}'^-, \theta_{sp}'^-)$ thus obtained, we construct a line $l^-$ representing the relationship between $\theta_{se}^-$ and $\theta_{sp}^-$. A similar approach is used to find a line $l^+$ that represents the relationship between the upper bounds $\theta_{se}^+$ and $\theta_{sp}^+$ for the two rates. We note that our procedure requires solving just a single polynomial equation, which is feasible in general [12]. For larger CPTs, however, the procedure becomes computationally more demanding, since a larger sample of points is required to assure good results.

We now have that, if the probability of interest $p(x_r^n \mid e_n)$ falls below $p^-$, the decision to withhold treatment remains unaltered for any pair $(\theta_{se}, \theta_{sp})$ below $l^-$. If $p(x_r^n \mid e_n) > p^+$, the decision to start treatment remains unaltered for any pair $(\theta_{se}, \theta_{sp})$ above $l^+$. Finally, if $p^- \leq p(x_r^n \mid e_n) \leq p^+$, the decision will be the same for any pair $(\theta_{se}, \theta_{sp})$ between $l^-$ and $l^+$.

Figure 6 now illustrates the sensitivity analysis. With our approximation procedure, two lines are established that serve to divide the unit square into three areas in which different decisions apply. For our example patient, we have that $p(vap_5 \mid e_5) = 0.9842 > p^+$. Since the original values for the sensitivity and specificity rates under study are $0.9$ and $0.95$ respectively, we conclude that the decision to start treatment right away is quite robust with regard to the CPT. The three bullets in the figure highlight three other interesting cases. For the bullet with $\theta_{sp} = 0.6$, we observe that the decision to test is only moderately robust since a small change in $\theta_{sp}$ or $\theta_{se}$ can alter the decision. For the bullet with $\theta_{sp} = 0.8$, the decision to treat is quite robust since only a major change in $\theta_{sp}$ or $\theta_{se}$ can induce another decision. Finally, for the bullet with $\theta_{sp} = 0.87$, the decision to test is not robust at all since a small change in $\theta_{sp}$ or $\theta_{se}$ suffices to alter the decision.

To conclude, we note that when the function $\widetilde{g}^{-1}$ is not invertible, our sampling procedure will result in multiple solutions. The unit square for $\theta_{se}$ and $\theta_{sp}$ will then be divided in compound areas per decision, similar to the compound intervals in the single-parameter case.

## 6 An approximate scheme

The number of constants in the sensitivity functions of a DBN and the number of propagations required to compute these constants grows linearly with $n$. Moreover, the computational burden of solving polynomials of high order can grow dramatically [12]. For a larger time scope, therefore, sensitivity analysis in view of threshold decision making can become quite hard. To reduce the order of the polynomials in the sensitivity functions and thereby the runtime requirements, we present a method for approximate sensitivity analysis that builds on the concept of *contraction* of a Markov process [1]. We discuss our method for DBNs with a single hidden process and review its extension to DBNs with multiple processes.

We consider two probability distributions $\mu$ and $\mu'$ over a variable $W$. Conditioning on a set of observations is known to never increase the relative entropy of these distributions. Denoting the conditioning by $o(\cdot)$, we thus have that

$$D[o(\mu) \| o(\mu')] \leq D[\mu \| \mu'] \qquad (2)$$

where $D$ stands for the relative entropy. Now, consider the extreme case where $\mu$ and $\mu'$ have their entire probability mass on two different states $w_i$ and $w_k$ respectively. We denote by $A(\cdot)$ the distribution that results from processing through the transition matrix $A$. Even though $\mu$ and $\mu'$ do not agree on any state, processing will cause them to place some mass on some state $w_j$. They then agree for a mass of $min[A(\mu(w_j; w_i)), A(\mu'(w_j; w_k))]$ on that state $w_j$. Based on this property, the *minimal mixing rate* of the matrix $A$ is defined as

$$\delta_A = \min_{i,k} \sum_j \min\left[A(\mu(w_j; w_i)), A(\mu'(w_j; w_k))\right]$$

Given the minimal mixing rate of a transition matrix $A$, the following theorem now holds [1]:

$$D[A(\mu)\|A(\mu')] \leq (1 - \delta_A) \cdot D[\mu\|\mu']$$

We say that the stochastic process with transition matrix $A$ *contracts* with probability $\delta_A$. Combining equation (2) with the previous theorem gives

$$D[A(o(\mu))\|A(o(\mu'))] \leq (1 - \delta_A) \cdot D[\mu\|\mu']$$

Performing conditioning on two different distributions and subsequently transitioning them, will thus result in two new distributions whose distance in terms of relative entropy is reduced by a factor smaller than one.

Our approximate method for sensitivity analysis now builds on the contraction property reviewed above. Suppose that we are interested in the probability of some state of the hidden variable $X_n$ at time step $n$. After entering the available evidence $e_n$ into the model, we can compute the exact posterior distribution $p(X_n \mid e_n)$. Building on the contraction property, however, we can also compute an approximate distribution $\widetilde{p}(X_n \mid e_n)$ starting from time step $n_\phi$, with $1 < n_\phi < n$, without losing too much accuracy. We define the *backward acceptable window* $\omega_{n,\epsilon}^\phi$ for time step $n$ with a specified level of accuracy $\epsilon$, to be the number of time steps we need to use from the past to compute the probability distribution of the hidden variable at time step $n$ within an accuracy of $\epsilon$. We now propose to perform sensitivity analysis for time step $n$ considering only the backward acceptable window $\omega_{n,\epsilon}^\phi$. Note that the resulting functions then include polynomials of order $O(n - n_\phi)$ rather than of order $O(n)$ compared to the true functions.

For a given level of accuracy $\epsilon$, we have to determine the maximum value of $n_\phi$ for which

$$D[p(X_n \mid e_n)\|\widetilde{p}(X_n \mid e_n)] \leq$$
$$(1 - \delta_A)^{n-n_\phi} \cdot D[p(X_{n_\phi} \mid e_{n_\phi})\|p(X_1)] \leq \epsilon$$

where $\widetilde{p}(X_n \mid e_n)$ denotes the approximate distribution of $X_n$ that is computed using $\omega_{n,\epsilon}^\phi$. Solving for $n_\phi$, we find that

$$n_\phi = \max\left\{1, n - \left\lfloor \frac{\log\left(\epsilon/D[p(X_{n_\phi} \mid e_{n_\phi})\|p(X_1)]\right)}{\log(1 - \delta_A)} \right\rfloor\right\} \quad (3)$$

where $\lfloor \cdot \rfloor$ stands for the integer part. Starting from $n_\phi = n$ and decreasing the value of $n_\phi$ one step at a time, we can readily establish the value of $n_\phi$ that first satisfies the equation (3). To this end, the interface algorithm needs to have computed and stored the exact posterior distributions $p(X_{n_o} \mid e_{n_o})$ for all $n_o \leq n$, given evidence $e_{n_o}$.

The procedure to compute the optimal value $n_\phi$ requires at most $n$ computations of (3) and thus is not very demanding from a computational point of view. We recall, however, that for the computation of $n_\phi$, the interface algorithm needs to have established the exact posterior distributions given the available evidence. Now in a full sensitivity analysis, the effects of parameter variation are being studied for a number of evidence profiles. The above procedure may then become rather demanding since for every such profile a full propagation with the interface algorithm is required. An alternative way would then be to approximate $n_\phi$ given $\epsilon$ from the start and perform the entire analysis with the backward acceptable window $\omega_{n,\epsilon}^\phi$. If we assume that $D[p(X_{n_\phi})\|p(X_1)]$ is bounded from above by a known constant $M$, we find that an approximate value for $n_\phi$ would satisfy

$$n_\phi \approx \max\left\{1, n - \left\lfloor \frac{\log(\epsilon/M)}{\log(1 - \delta_A)} \right\rfloor\right\}$$

Note that for a given $\epsilon$ and $\delta_A$, the higher the value of $M$, the smaller the value of $n_\phi$ and hence the larger the backward acceptable window. Knowledge of the domain under study can help in determining a suitable value for $M$. For a patient profile for example, $M$ can be determined by inserting *worst-case scenario* observations for the first time step and computing for that time the posterior probability distribution for the hidden variable from which $M$ can be readily established. The complexity that our method now entails is just the complexity of computing $M$ which is similar to performing a single propagation for a single time step. This computational burden is considerably less than the burden of performing $n_\phi$ time steps of exact inference, which we thereby forestall in the sensitivity analysis. Note that for some patients the computation of $n_\phi$ based upon this value $M$ will lead to a larger backward acceptable window than the one computed directly from equation (3).

In view of sensitivity analysis, we observe that the value of $n_\phi$ that is established as outlined above, is based on the original values of all parameters of the model under study. We further observe that the minimal mixing rate $\delta_A$ used in the computation of $n_\phi$ is algebraically dependent only on the model's transition parameters. Using $\omega_{n,\epsilon}^\phi$ based upon $n_\phi$ for sensitivity analysis, therefore, is guaranteed to result in approximate sensitivity functions within an accuracy of $\epsilon$ for any non-transition parameter. For transition parameters, this guarantee does not hold in general. We note, however, that for the original value of a transition parameter, the difference between the true probability of interest and the approximate one is certain to be smaller than $\epsilon$. Since the value $n_\phi$ changes with $\delta_A$ in a stepwise manner only, this property holds for a range of values for the parameter. Figure 7 illustrates the relationship between $n_\phi$ and $\delta_A$ given particular values for $n$, $\epsilon$ and $M$. We observe from the figure that there is a range of values of $\delta_A$ for which the value of $n_\phi$ stays the same. We expect a similar property to hold for a range of values for the transition parameter $\theta_a$. We are currently studying this issue and hope to report results in the near future.

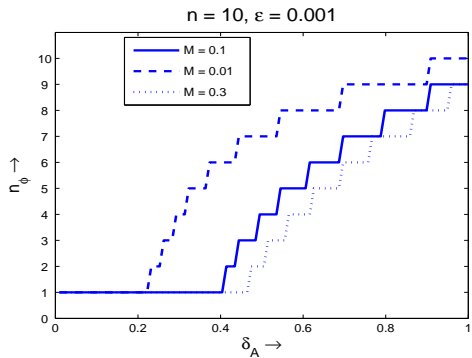

Figure 7: The relation between $n_\phi$ and $\delta_A$.

In general, a DBN with multiple interacting subprocesses can be represented as a single-process stochastic model with a *global* transition matrix $A_G$ by enumerating all combinations of values for the subprocesses. We can show that a lower bound on the global mixing rate $\delta_{A_G}$ can be computed from knowledge of the contraction rates of the individual subprocesses of the model [6]. For a DBN composed of several sparsely interacting subprocesses each of which is fairly stochastic, we expect a reasonable lower bound on the mixing rate $\delta_{A_G}$. We recall that the larger the mixing rate, the larger $n_\phi$ and the smaller the backward window that we can acceptably use for the sensitivity analysis. For all patients in our application for ICU care for example, we found that instead of using the observations for 10 days upon performing a sensitivity analysis for the probability of VAP, we could use the observations from day 5 on with an average error smaller than $\epsilon = 0.003$. This result is quite promising as it shows that even if the dynamic processes of a DBN are only moderately stochastic, the backward acceptable window can still be small enough to allow for good approximations of the sensitivity functions.

## 7 Conclusions

In this paper we presented functional forms to express the sensitivity of a probability of interest of a DBN in terms of one or more parameters from a single CPT. We used these sensitivity functions for studying the robustness of the output of the network in view of the threshold decision making. In addition, we presented an approximate scheme for computing the constants involved in the sensitivity functions that is less demanding than an exact algorithm yet incurs only a small loss in accuracy. We illustrated our results with a real-life dynamic network for ICU care.

## Acknowledgements

This research was (partly) supported by the Netherlands Organization for Scientific Research (NWO).